\pdfoutput=1

\documentclass[11pt]{article}

\usepackage[final]{acl}

\usepackage{times}
\usepackage{latexsym}

\usepackage[T1]{fontenc}

\usepackage[utf8]{inputenc}

\usepackage{microtype}

\usepackage{inconsolata}

\usepackage{graphicx}

\usepackage{arydshln}
\usepackage{pifont}
\newcommand{\cmark}{\ding{51}}%
\newcommand{\xmark}{\ding{55}}%

\usepackage{amsmath}

\title{Summarizing Long Regulatory Documents with a Multi-Step Pipeline}

\author{Mika Sie \\
  Utrecht University \\ 
  \texttt{mikasie6@gmail.com} \And
  Ruby Beek  \\
  Power2X \\
  \texttt{ruby.beek@power2x.com} \And
  Michiel Bots \\
  Power2X  \\
  \texttt{michiel.bots@power2x.com} \AND
  Sjaak Brinkkemper \\
  Utrecht University \\
  \texttt{s.brinkkemper@uu.nl} \And
  Albert Gatt \\
  Utrecht University\\
  \texttt{a.gatt@uu.nl}}

\begin{document}
\maketitle
\begin{abstract}
Due to their length and complexity, long regulatory texts are challenging to summarize. To address this, a multi-step extractive-abstractive architecture is proposed to handle lengthy regulatory documents more effectively.
In this paper, we show that the effectiveness of a two-step architecture for summarizing long regulatory texts varies significantly depending on the model used. Specifically, the two-step architecture improves the performance of decoder-only models. For abstractive encoder-decoder models with short context lengths, the effectiveness of an extractive step varies, whereas for long-context encoder-decoder models, the extractive step worsens their performance. This research also highlights the challenges of evaluating generated texts, as evidenced by the differing results from human and automated evaluations. Most notably, human evaluations favoured language models pretrained on legal text, while automated metrics rank general-purpose language models higher.
The results underscore the importance of selecting the appropriate summarization strategy based on model architecture and context length. 
\end{abstract}

\section{Introduction}
\label{sec: Introduction}
Automatic text summarisation (ATS) involves generating a compressed, concise, and fluent version of an input text while preserving its main key points. A summary proves useful because it helps people process and understand texts faster and better. Summarizing regulatory texts is important for making complex legal language more accessible and ensuring compliance by condensing information into a concise, understandable format.
\footnote{Code and models are available on \href{https://github.com/MikaSie}{GitHub} and \href{https://huggingface.co/MikaSie}{HuggingFace}.}

Current ATS methods use either extractive or abstractive summarization. 
An advantage of extractive summarization is that it captures sentences and information literally, resulting in a factually consistent summary. However, the summary is harder to read and less intuitive as sentences are copied and combined. Abstractive summaries are more coherent and fluent as they summarize texts in a human-like fashion. But it also has disadvantages because an intricate understanding of the original text is required and the summary can be factually inconsistent. In this paper, our aim is to explore the advantages of both strategies, as we leverage them for the summarisation of very lengthy, regulatory documents.

A regulatory text is a formal document issued by a government or regulatory body that outlines rules, guidelines, or standards to govern the conduct, practices, or operations within a specific industry, sector, or jurisdiction. Regulatory documents are difficult to process due to their extensive size, unique structure, numerous citations and references, ambiguity, and domain-specific vocabulary.  
Current automatic summarization tools face challenges with regulatory texts, either because their length exceeds the context length of LLMs, or because the length and structure of the input document raise the risk of omissions in the summary. Leaving out important information 
could have major negative effects. 

\begin{figure}
    \centering
    \includegraphics[width= 0.48\textwidth]{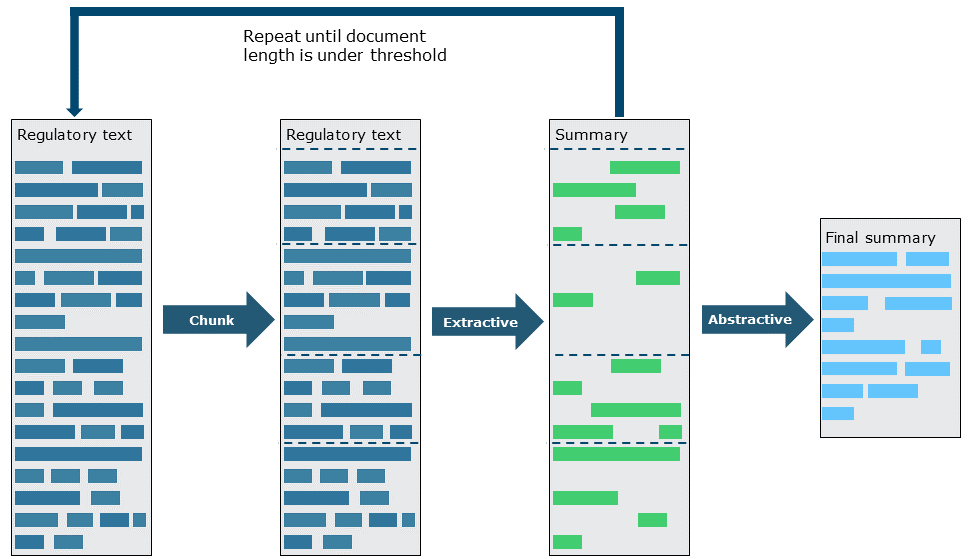}
    \caption{Summarization process proposed by this research. Dotted lines indicate the borders of the chunks.}
    \label{fig: summarization process}
\end{figure}

This paper compares two-step and multi-step summarisation methods for regulatory documents, comparing the effectiveness of different neural model architectures and combinations. Our approach consists of the following steps, illustrated in Figure \ref{fig: summarization process}.
the document is segmented into smaller or `chunks'. Each chunk is then processed by an extractive summarization model, and all resulting summaries are concatenated. This extractive step may need to be conducted iteratively.
The outcome of extraction is then summarized in an abstractive manner, creating a final summary. Combining these two summarization steps could prove useful in handling the large size of the original text. It uses extracted salient sentences to develop a coherent, fluent summary. 
Similar architectures have been used on different types of texts and have shown promising results \cite{Pilault2020, Zhang2022, Klaus2022, Bleiweiss2023}. However, summarizing long regulatory documents using this architecture has been researched less extensively. In particular, our goal is to evaluate various models used for each step to identify the most effective combination of models for the summarization task, paying particular attention to whether preliminary extraction is more beneficial if performed with domain-specific (legal) rather than domain-general models. A second important goal is to compare the effect of context length on the quality of the generated summaries: models allowing longer context lengths need less extraction. Given the growing trend for large language models to allow longer document lengths, it is increasingly important to understand whether such models are able to acquire a comprehensive understanding of a full document, or whether preliminary distillation of information is helpful \cite{li_loogle_2023}.

\section{Related work}
\label{sec: Related work}

\paragraph{Long document summarisation}
Pretrained language models (LMs) struggle with long texts due to limitations on input context length. For example, BERT \cite{Devlin2019} and  T5 \cite{Raffel2020} have a context length of 512 tokens while PEGASUS's \cite{Zhang2020Pegasus} and BART's \cite{Lewis2020} context length is 1024 tokens. 

To counter this limitation, some long document architectures incorporate a different self-attention mechanism, 
calculating attention between specific parts of the sequence instead of calculating the attention for every possible combination of the sequence. This enables them to process long sequences because the computation requirements will not grow quadratically. 
Longformer \cite{beltagy2020longformer} is an encoder-only architecture based on RoBERTa \cite{liu2019roberta}, designed to handle long-range dependencies more efficiently than standard transformers, and accepting inputs of up to 4096 tokens. 
It employs a combination of global attention and sliding window attention instead of full attention, which scales linearly with the input sequence. LED \cite{beltagy2020longformer} adds a decoder to the Longformer architecture, turning it into the Longformer Encoder-Decoder model. The decoder does use the full attention mechanism but LED retains its linear computation capability. Similar examples of LMs designed for longer documents include BigBird \cite[which accepts a context length of 4096 tokens;][]{Zaheer2020}, LongT5~\cite{Guo2022} and PegasusX~\cite{phang-etal-2023-investigating}, both of which accept contexts of 16,384 tokens.

Extending context length is often a goal in recent releases of decoder-only LLMs, such as the GPT family of models. Other examples include \href{https://huggingface.co/togethercomputer/LLaMA-2-7B-32K}{LLaMA-2-7B-32k} \cite{TogetherAI}, which is an LLM based on LLaMA-2 \cite{touvron2023llama} with a context length of 32768.



\paragraph{Multi-step summarisation}
The idea of multi-step methods is to leverage both extractive and abstractive techniques to alleviate the burden of summarising very long documents.
Pilault et al. \cite{Pilault2020} 
add one extractive step before generating the abstractive summary. The extractive parts are then used beside the original text as input for the transformer. 
A related approach is taken in CreativeSumm \cite{Kim2022} for the summarisation of lengthy movie scripts. \citet{Liu2018} summarise Wikipedia articles by first performing an extractive step, using the extracted sentences as additional input to the summariser. \citet{Bleiweiss2023} propose a two-step method for long biographical novels. \citet{Klaus2022} make use of a two-step method to summarize legal regulatory documents. Klaus et al. use TextRank \cite{Mihalcea2004}, a graph-based extractive summarization approach, for the first extractive step and BERT \cite{Devlin2019} or RoBERTa \cite{liu2019roberta} for a second extractive step. 

A generalisation of the two-step strategy was proposed in the form of Summ\textsuperscript{N} \cite{Zhang2022}. Summ\textsuperscript{N} splits the data samples and generates coarse summaries, possibly over multiple stages ($N$), before producing a final fine-grained abstractive summary. This method outperformed previous state-of-the-art methods on different datasets. Different from our work, Summ$^N$ makes use of abstractive summarisation for both the coarse-grained and the final, more fine-grained summarisation steps. Instead, we use extractive summarisation for the first stage.

Inspired by multi-step methods, we experiment in this paper with various combinations of extractive and abstractive steps, in an effort to identify the best architecture for summarisation of long, regulatory documents. 

\paragraph{Divide-and-conquer (chunking) strategies}
An interesting class of approaches to long document summarisation involves a `divide-and-conquer' strategy. Briefly, the idea is to chunk the document into sub-parts before summarisation, where sub-part identification may also exploit the document structure. Examples of this are the context-aware chunking strategy for academic articles used in DANCER \cite{Gidiotis2020} and the work of \citet{Shen2022}, whose model directly learns the correspondence between document sections and summary parts. In our work, we also explore the role of chunking strategies and their effectiveness in producing coherent summaries.

\paragraph{Domain-specific Legal Language Models}
An important question in the processing of texts in specialised domains is whether in-domain pretraining is beneficial, given that specialised domains have stylistic and other peculiarities. Relevant to the present paper is the case of legal text (of which regulatory texts are a subset), which has well-studied distinctive stylistic characteristics~\cite{Turtle1995,Kanapala2019,Jain2021}. 
Studies have shown that in-domain pretraining can be beneficial in downstream NLP tasks \cite{Gururangan2020} and domain-specific LMs have been developed for healthcare \cite{huang2020clinicalbert, Lee2020}, science \cite{Beltagy2019} and finance \cite{yang2020finbert,wu2023bloomberggpt}, among many others. Pre-trained LMs for law include Lawyer 
LLaMA\cite{huang2023lawyer}, Lawformer \cite{Xiao2021}, LegalLongformer \cite{Mamakas2022}, PEGASUS-Billsum \cite{Zhang2020Pegasus}, LegalBERT \cite{Chalkidis2020}, CaseLawBERT \cite{Zheng2021}, PoL-BERT \cite{Henderson2022} and LexLM \cite{Chalkidis2023}. In an early study, \citet{Chalkidis2020} showed that LegalBERT consistently outperformed BERT-based models on a variety of NLP tasks, including EURLEX57K \cite{Chalkidis2020-1}, ECHR-CASES \cite{Chalkidis2020-2}, and CONTRACTS-NER \cite{Chalkidis2017}.
Building on this work, \citet{Mamakas2022} introduced LegalLongformer, initialised with LegalBERT's parameters, to handle long legal texts. \citet{Chalkidis2023} introduced LexLM, a model pre-trained on a multinational English legal data. Additionally, they introduced a version of LexLM utilizing the Longformer \cite{beltagy2020longformer} attention mechanism, enhancing the capability to handle long legal documents.
In comparative evaluations, LexLM models outperformed other legal LMs, such as CaseLawBERT and PoL-BERT, particularly in prior knowledge assessment and downstream task performance. Notably, RoBERTa \cite{liu2019roberta} also showed strong performance, occasionally surpassing some specialized legal models. 

Building on these observations, in our experiments we also compare general-purpose models with a representative subset of legal LMs, particularly for the extractive summarisation step.

\section{Method}
\label{sec: Method}
\begin{figure}[!t]
    \centering
    \includegraphics[scale=0.23]{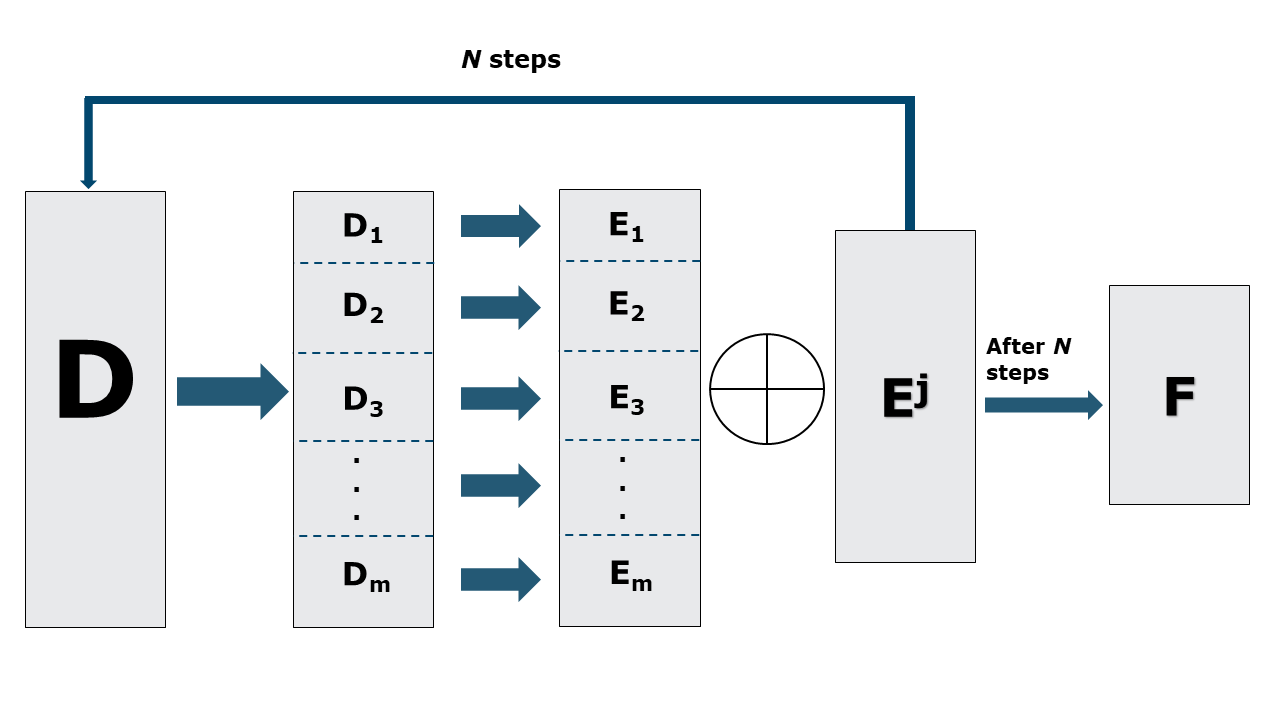}
    \caption{Visualisation of the summarisation process. $N$ represents the amount of extractive steps and the $\oplus$ symbol represents the concatenation process.}
    \label{fig:process}
\end{figure}

Our approach to the summarisation of long regulatory documents is a multi-step process consisting of extraction followed by abstraction, where extraction is intended to alleviate the problem of limited context length accepted by a model. In particular, if the length of a source document $\vert D \vert$ exceeds the context length $K$ of an abstractive model, creating an intermediate extractive summary could help identify essential information across the document span, a more informed strategy than truncating the document to fit within $K$.

The overall process is visualised in Figure~\ref{fig:process}. We view a source document $D = \{D_1, D_2, D_3, ... D_m\}$ as a sequence of chunks $D_i$.
A chunk is summarised by an extractive summarisation model, which produces an intermediary summary $E^j = E^j_1 \oplus E^j_2 \oplus E^j_3 \oplus, ..., \oplus E^j_M$, where $E^j_i$ represents an intermediate summary of chunk $D_i$ at extractive step $j$. Thus, the intermediate summary $E^j$ comprises the summaries of all the chunks concatenated in the same order as in the original text. The extractive summarisation model has a compression ratio $R \in [0,1]$. One way to define $R$ is in terms of the ration of the length of an article and that of its summary~\cite{Grusky2018}; below, we also explore other possible definitions for $R$. Before the summarisation is performed, the number of extractive steps $N$ taken is determined, such that the extractive summary produced at step $n \le N$ is the input to the extractive step $n+1 \leq N$. The extractive summary after $N$ steps is the input to the abstractive summarisation model, which yields the final summary $F$. 
\subsection{Dataset}
\label{subsec: Dataset}

The dataset used to fine-tune the abstractive model is EUR-Lex-Sum \cite{Aumiller2022}. This dataset consists of documents from the European Union law platform with corresponding manually curated summaries. Only the English part of the dataset, composed of 1504 document -summary pairs, was used for this task. It has been divided into training, validation, and test sets, containing 1129 pairs, 187 and 188 pairs, respectively. 
The dataset is characterised by a small number of documents whose length far exceeds that of the others. To ensure consistency in our evaluation, we define any document whose word count is more than two standard deviations above the mean as an outlier and remove it from the training, validation and test subsets originally provided by \citet{Aumiller2022}. In total, 62 instances were removed by this criterion. The final dataset consists of 1091 training, 172 validation and 179 test samples.

\subsection{Architecture}
\label{subsec: Architecture}

\paragraph{Extractive step(s)}
As described above, documents are first summarised over $N$ extractive steps. Note that the extractive step is only performed if the length of the document exceeds the context length $K$ of the abstractive model. The number of extractive steps needed ultimately depends on the compression ratio $R$ that we require for the summarisation, corresponding to two-step (one extraction step followed by abstraction) and multi-step approaches. We experiment with three different strategies for computing $R$ for an abstractive model with context length $K$and a document of length $\vert D \vert$. Note that $K$ and $\vert D \vert$ are fixed in advance for a given model and document. 

Our first strategy is to use a {\bf fixed compression ratio}, empirically setting $R=0.4$. In this case, $N \geq 1$ and is estimated as follows (see Appendix \ref{appendix: Derivation of N} for details of how this is derived):

\begin{equation}
\label{equation: calculate steps for fixed ratio}
N = \left\lceil \frac{\log\left(\frac{K}{|D|}\right)}{\log(R)} \right\rceil
\end{equation}

The second strategy is to use a {\bf dependent compression ratio}, which depends on the document's size and the abstractive model's context length, resulting in $N=1$:

\begin{equation}
\label{equation: dependent ratio}
R = \frac{K}{|D|}
\end{equation}

The final strategy is a {\bf hybrid ratio}, where we perform $N-1$ extractive steps with a fixed ratio, with a final extractive step $N$ using a dependent ratio. The hybrid ratio could be more effective than the fixed ratio because it is focused on ensuring that the final intermediate summary optimally fits the context length of the abstractive model. We define the hybrid ratio as follows:

\begin{equation}
R =
\begin{cases}
    0.4 & \text{for steps } 1, 2, \ldots, N-1 \\
    \frac{K}{|D|} & \text{for step } N
\end{cases}
\end{equation}

\paragraph{Extractive models}
One of our goals is to compare the impact of domain-specific LMs and general-purpose LMs. In what follows, non-domain-specific LMs will be referred to as 'general' LMs, and domain-specific legal LMs will be referred to as 'legal' LMs.
The top panel of Table \ref{tab: Summarisation models used} lists all the extractive summarisation models used. Based on this comparison, we aim to identify the optimal extractive model.

\begin{table*}[h]
\small
\centering
\hspace*{-1cm}\begin{tabular}{lllll}
\hline
Model & Context length & Legal LM & Type & Architecture \\
\hline
RoBERTa \cite{liu2019roberta} & 512 & \xmark & Extractive & Encoder \\
Longformer \cite{beltagy2020longformer} & 4096 & \xmark & Extractive & Encoder \\
LegalBERT-SC \cite{Chalkidis2020} & 512 & \cmark & Extractive & Encoder \\
LexLM \cite{Chalkidis2023} & 512 & \cmark & Extractive & Encoder \\
LexLM - Longformer \cite{Chalkidis2023} & 4096 & \cmark & Extractive & Encoder \\
\hdashline
BART \cite{Lewis2020} & 1024 & \xmark & Abstractive & Encoder-Decoder \\
T5 \cite{Raffel2020} & 512 & \xmark & Abstractive & Encoder-Decoder \\
LongT5 \cite{Guo2022} & 16384 & \xmark & Abstractive & Encoder-Decoder \\
Pegasus \cite{Zhang2020} & 1024 & \xmark & Abstractive & Encoder-Decoder \\
PegasusX \cite{phang2022investigating} & 16384 & \xmark & Abstractive & Encoder-Decoder \\
Llama3 \cite{meta2024} & 8192 & \xmark & Abstractive & Decoder \\
\hline
\end{tabular}
\caption{Summarisation models used. The context length is expressed in number of tokens. Top: models used for extractive summarisation; bottom: models used for abstractive summarisation.}
\label{tab: Summarisation models used}
\end{table*}

We compare all the extractive models with the three ratio types described above, with a view to determining the optimal extractive strategy to support abstractive summarisation. To identify the optimal extractive model, we compare the impact of different extractive models and compression ratios on downstream abstractive summarisation with BART~\cite{Lewis2020}. Specifically, we compare the output of a BART summariser, finetuned on using input from different extractive models. We compare this to a baseline BART model with no extractive steps. In total, we compare sixteen model configurations. The optimal extractive strategy under this experimental setting was then used to fine-tune subsequent abstractive models.

\paragraph{Abstractive step}
The abstractive step was only performed once the length of the intermediate summary $|E^j|$ is within the context length $K$ of an abstractive summarisation model. The abstractive step involves creating the final summary $F$ by an abstractive summarisation model fine-tuned on the intermediate summary $E^j$. 

We compare a variety of abstractive models, listed in the bottom panel of Table~\ref{tab: Summarisation models used}. The context length of the abstractive summarisation model is an important consideration as it affects the number of extractive steps. A longer context length implies that fewer extractive steps need to be taken. By hypothesis, the quality of the final summary should be higher the fewer the extractive steps, since there is less potential in this case for information loss. To quantify this, we chose models that permit a direct comparison of context length effects, while keeping architecture largely constant. We compare T5 \cite{Raffel2020} against LongT5 \cite{Guo2022}, and Pegasus \cite{Zhang2020} against PegasusX \cite{phang2022investigating} to determine the effect of a long context length in the abstractive summarization model. Finally, we include Llama3~\cite{meta2024}, as an example of a SOTA large language model based on a decoder architecture (T5 and Pegasus are encoder-decoder models).

Full parameter fine-tuning was performed for all abstractive models except Llama3, which was fine-tuned using QLoRA \cite{dettmers2023qlora} as full parameter fine-tuning was not feasible due to its size. Data had to be prepared in a different way for Llama3 as it is the sole decoder-only model used in our experiments. A single combined sequence is used instead of separate input and output sequences. To accommodate a summary of 1500 tokens, 1500 tokens are subtracted from the model’s context length, resulting in an effective context length of 6692 tokens for Llama3. The extractive summarisation process was adjusted to summarise the reference text to fit within this 6692-token limit, ensuring minimal truncation. 
See Appendix \ref{appendix: Hyperparameter settings} and \ref{appendix: Llama3 hyperparameter settings} for more details on model finetuning, including hyperparameters.

\subsection{Evaluation}
\label{subsec: Evaluation}
\paragraph{Evaluation metrics} 
Multiple evaluation metrics were used to assess the proposed architecture from different aspects. This research employed ROUGE-1, ROUGE-2, ROUGE-L \cite{Lin2004}, BERTScore \cite{Zhang2020}, BARTScore \cite{Yuan2021}, and BLANC \cite{Vasilyev2020}. Details of the implementations used for the evaluation metrics are in Appendix~\ref{appendix: Evaluation metrics details}.

\begin{table*}[!h]
  \small
  \hspace*{-1.5cm}
    \centering
  \begin{tabular}{lp{8cm}}
  \hline
    Criterion & Description \\
    \hline
    Factual Correctness & Evaluation of how factually correct the summary is relative to the source document. \\
    Usability & Assessment of how practical and user-friendly the summary is. \\
    Accuracy & Assessment of the precision and correctness of the information in the summary. \\
    Fluency & Assessment of the summary's smoothness and ease of reading in terms of form, content, and grammar. \\
    Coherence & Measure of how logical the summary is to it is linguistic context.\\
    \hline
  \end{tabular}
  \caption{Criteria for human evaluation.}
  \label{tab: Criteria for human evaluation.}
\end{table*}

\paragraph{Expert evaluation} Besides automated metrics, we also performed a small-scale qualitative human evaluation involving expert readers. The human evaluation provides insights into the quality of the summaries, complementing the quantitative data from automated metrics with qualitative feedback. 
After selecting the optimal extractive model and training the abstractive models, we generate summaries of a new text which is not in the training dataset.\footnote{The text in question is the \href{https://eur-lex.europa.eu/legal-content/EN/TXT/PDF/?uri=CELEX:32023R0956}{Carbon Border Adjust Mechanism document} \cite{EU2023Regulation}.} Summaries generated with the different abstractive models were compared by the expert readers. This document was chosen specifically because the expert readers were already familiar with the contents and, hence, were able to judge summary quality more reliably. 

The evaluators were two experts from the company {\tt ANON}, a collaborator on this project whose personnel have extensive experience with regulatory documents issued by the European Union. 
The experts were asked to read summaries generated by different summarization architectures and evaluate them based on a set of criteria. The criteria included \textit{Factual Correctness}, \textit{Usability}, \textit{Accuracy}, \textit{Fluency}, and \textit{Coherence}. Each criterion was rated on a scale from 1 to 5. Detailed descriptions of these criteria can be found in Table \ref{tab: Criteria for human evaluation.} and are based on the findings of \citet{Howcroft2020}'s meta-review of constructs used in human evaluation of Natural Language Generation systems. In addition to scoring the summaries, experts were also asked to comment on the quality of summaries.

Due to resource and time constraints, we selected specific architectures to be included in the qualitative evaluation. To analyse the impact of different extractive models, we compare different versions of BART, using 
(1) the best extractive model; (2) no extractive step; (3) the best legal LM for extraction; and (4) the best long-context extractive model. To analyse the impact of different abstractive strategies, we also include (5) the best long-context abstractive model; and (6) the best decoder-only model.

\section{Results}
\label{sec: Results}
\begin{table*}[!t]
    \small
    \centering
    \hspace*{-1cm}\begin{tabular}{llllccccc}
        \hline
        Extractive model & Ratio type & R1 & R2 & RL & BERTScore & BARTScore & BLANC \\
        \hline
        N/A & No extraction & 0.4590 & 0.1954 & 0.2174 & 0.8702 & \textbf{-3.4154} & 0.1029 \\
        \hdashline
        RoBERTa & Fixed & 0.4670 & 0.1798 & 0.2171 & 0.8692 & -3.5654 & 0.1040 \\
        RoBERTa & Dependent & \textbf{0.4873} & \textbf{0.1974} & \textbf{0.2247} & \textbf{0.8721} & -3.5590 & 0.1272 \\
        RoBERTa & Hybrid & 0.4809 & 0.1889 & 0.2193 & 0.8700 & -3.5781 & \textbf{0.1296} \\
        \hdashline
        LegalBERT & Fixed & 0.4390 & 0.1766 & 0.2158 & 0.8700 & -3.4893 & 0.1099 \\
        LegalBERT & Dependent & 0.4619 & 0.1854 & 0.2174 & 0.8713 & -3.5143 & 0.1117 \\
        LegalBERT & Hybrid & 0.4469 & 0.1774 & 0.2137 & 0.8665 & -3.5714 & 0.1098 \\
        \hdashline
        LexLM & Fixed & 0.4571 & 0.1745 & 0.2123 & 0.8692 & -3.6130 & 0.1154 \\
        LexLM & Dependent & 0.4859 & 0.1954 & 0.2227 & 0.8713 & -3.5441 & 0.1277 \\
        LexLM & Hybrid & 0.4582 & 0.1792 & 0.2135 & 0.8665 & -3.5639 & 0.1102 \\
        \hdashline
        Longformer & Fixed & 0.4436 & 0.1686 & 0.2103 & 0.8684 & -3.5901 & 0.1029 \\
        Longformer & Dependent & 0.4613 & 0.1874 & 0.2194 & 0.8712 & -3.5835 & 0.1238 \\
        Longformer & Hybrid & 0.4778 & 0.1862 & 0.2181 & 0.8703 & -3.5697 & 0.1256 \\
        \hdashline
        LexLM-Longformer & Fixed & 0.4250 & 0.1584 & 0.2041 & 0.8659 & -3.6141 & 0.0959 \\
        LexLM-Longformer & Dependent & 0.4751 & 0.1852 & 0.2164 & 0.8689 & -3.5344 & 0.1272 \\
        LexLM-Longformer & Hybrid & 0.4619 & 0.1819 & 0.2189 & 0.8692 & -3.5833 & 0.1199 \\
        \hline
    \end{tabular}
    \caption{\label{tab: Results for all extractive summarization models}
    Results for all extractive summarization models in combination with BART.}
\end{table*}

\begin{table*}[!h]
\centering
\small
\hspace*{-1cm}\begin{tabular}{lllcccccc}
\hline
Abstractive model &    Ratio type &  R1 &  R2 &  RL &  BERTScore &  BARTScore &  BLANC \\
\hline
BART & No extraction & 0.4590 & 0.1954 & 0.2174 & 0.8702 & -3.4154 & 0.1029 \\
BART & Dependent         &   \textbf{0.4873} & \textbf{0.1974} & 0.2247 & \textbf{0.8721} & -3.5590 & 0.1272 \\
\hdashline
T5 & No extraction       & 0.3033 & 0.1241 & 0.1994 & 0.8443 & -2.1585 & 0.0760 \\
T5 & Dependent           & 0.2934 & 0.0926 & 0.1857 &  0.8404 & -2.2234 & 0.0812 \\
\hdashline
LongT5 & No extraction   & 0.3261 & 0.1309 & 0.2192 & 0.8497 & -2.2195 & 0.1128 \\
LongT5 &    Dependent    & 0.2854 & 0.0969 & 0.0969 & 0.8444 & -2.0423 & 0.1051 \\
\hdashline
Pegasus & No extraction  & 0.3305 & 0.1293 & 0.2260 & 0.8499 & \textbf{-1.8067} & 0.0923 \\
Pegasus &    Dependent   & 0.3067 & 0.0911 & 0.2021 & 0.8435 & -1.8940 & 0.0952 \\
\hdashline
PegasusX & No extraction & 0.3673 & 0.1622 & \textbf{0.2304} & 0.8523 & -2.4528 & 0.1086 \\
PegasusX &    Dependent  & 0.3052 & 0.1162 & 0.1960 & 0.8413 & -2.4305 & 0.0999 \\
\hdashline
Llama3 & No extraction &   0.4088 &   0.1816 &   0.2107 &     0.7854 &    -3.3424 & 0.1177 \\
Llama3 &    Dependent  &   0.4474 &   0.1885 &   0.2284 &     0.8687 &    -3.1268 & 0.1231 \\
\hline
\end{tabular}
\caption{Evaluation results of all abstractive models with and without an extractive step.}
\label{tab: Evaluation results of all abstractive models with and without an extractive step}
\centering
\end{table*}

\begin{table*}[!h]
\small
\centering
\hspace*{-2cm}
\begin{tabular}{llllccccc}
\hline
         & Extr. model & Ratio & Abstr. model & FC & U & Acc & Fl & Coh \\
\hline
1 & RoBERTa & Dep. & BART & 2.0 & 2.0 & 1.5 & 1.5 & 2.0 \\
2 & - & NE & BART & 3.5 & 1.0 & 2.0 & \textbf{3.0} & 1.5 \\
3 & LexLM & Dep. & BART & \textbf{4.0} & \textbf{3.5} & \textbf{3.0} & \textbf{3.0} & \textbf{3.0} \\
4 & Longformer & Dep. & BART & 2.0 & 2.0 & 2.5 & 1.5 & 2.0 \\
5 & - & NE & PegasusX & 3.5 & 1.0 & 2.5 & \textbf{3.0} & 1.0 \\
6 & RoBERTa & Dep. & Llama3 & 3.0 & 2.5 & 2.5 & 2.5 & 2.0 \\
\hline
\end{tabular}
\caption{Average human evaluation results. Dep: Dependent ratio; NE: No extraction; FC: Factual Correctness; U: Usability; Acc: Accuracy; Fl: Fluency; Coh: Coherence}
\label{tab: Average human evaluation results}
\end{table*}

\subsection{Comparison of extractive models}
\label{subsec: Comparison extractive models} 
Table \ref{tab: Results for all extractive summarization models} contains the results on different metrics for abstractive summarisation using BART, in combination with different extractive strategies. It can be seen that RoBERTa with a dependent ratio scores the highest on ROUGE-1, ROUGE-2, ROUGE-L, and BERTScore. RoBERTa with a hybrid ratio achieves the highest score on BLANC. On the other hand, the best BARTScore is obtained when we do not combine any extractive summarisation to compress the input to BART.

In the rest of this section, we discuss these results in light of the different experimental conditions.

\subsubsection{Effect of number of extractive stages}
\label{subsubsec: Effect of number of stages}

The results indicate that models using the dependent ratio type generally achieve higher performance across most metrics. Notably, the RoBERTa model with the dependent ratio type attains the highest scores in ROUGE-1 (0.4873), ROUGE-2 (0.1974), ROUGE-L (0.2247), and BERTScore (0.8721), suggesting superior performance in these areas. However, the BART model without any extractive steps achieves the best scores in BARTScore (-3.4154) and BLANC (0.1700), indicating a stronger performance in these specific metrics despite not utilizing extraction.

Using a multi-step architecture, that is, one that performs multiple extractive iterations (up to $N$; see Section~\ref{sec: Method}), sentences from differnt document chunks get combined during the summarization process. This could introduce noise and consequently fail to capture the most relevant and coherent information, resulting in lower performance. It seems that using a single extractive step is more effective at capturing the most important sentences out of a chunk relative to the context of the global document. We hypothesise that this explains the superiority of the dependent ratio (where $N=1$) on most metrics.

\paragraph{Effect of Legal Language Models}
General-purpose LMs such as RoBERTa achieve slightly higher scores across all metrics except BARTScore, compared to legal LMs. For this comparison, RoBERTa was compared against LegalBERT and LexLM, and Longformer was compared against LexLM-Longformer to accommodate for the context lengths. 

These results indicate that, when used as extractors for preliminary document compression, the broad range of training data types that general-purpose LMs are exposed to gives them an advantage in locating important information in the document. In contrast, legal LMs can suffer from a `narrow' focus, resulting in less coherent and comprehensive extractive summaries. This insight suggests that general LMs can be effective for domain-specific tasks, at least for preparatory steps such as the one considered here.

\paragraph{Effect of context length for the extractive step}
Models with shorter context lengths for the extractive step achieve higher scores across all metrics. RoBERTa was compared against Longformer for general LMs and LegalBERT and LexLM against LexLM-Longformer for legal LMs. This approach ensures a fair comparison by accommodating general and legal language model differences. 

This finding is surprising, since one would assume that longer-context models would perform better by capturing more global context. However, when sequences are excessively long, the models might struggle to maintain and encode all relevant information, leading to reduced sensitivity to portions of the input, in line with findings such as those reported by~\citet{fu_decoder-only_2023}, among others. 

This could explain why shorter context models, which deal with more manageable chunks of information, consistently perform better in the extraction task.

\paragraph{Optimal extractive model}
Based on Table \ref{tab: Results for all extractive summarization models}, RoBERTa with a dependent ratio will be chosen as the optimal extractive model and is used in the remainder of the experiments reported below.

\subsection{Comparison of abstractive models}
\label{subsec: Comparison abstractive models}
For every abstractive model, two versions are compared: one leveraging RoBERTa with a dependent ratio and one without using any extractive step at all. The results for all abstractive models and their variants can be seen in Table \ref{tab: Evaluation results of all abstractive models with and without an extractive step}. 
For clarity, models that incorporate an extractive step will be referred to by the name of the abstractive model. Models that do not use an extractive step will be denoted by appending “-NE” to the name of the abstractive model, where “NE” signifies “No Extraction”.

 
\paragraph{Effect of extractive step}
The performance of encoder-decoder abstractive summarization models generally worsens when using one extractive step, though this differs per model. This is evident in the results for T5, LongT5, Pegasus, and PegasusX, where the versions without extraction tend to outperform their counterparts with an extractive step. BART presents a more varied picture as it differs per metric in which variant scores higher. Since encoder-decoder models generate a condensed representation of the text, one explanation for these results is that by introducing an intermediate extractive summary we compromise the performance of the encoder. This could happen because the intermediate summary is less coherent than the input document as a whole.

LLama3, the decoder-only model seems to benefit from an additional extractive step, obtaining better results on all metrics when compared to the version with no extraction. The beneficial effect of extraction here is likely due to the limited context of Llama3 and the risk of loss of sensitivity to longer inputs, as decodig proceeds~\cite{fu_decoder-only_2023}. These shortcomings could be mitigated by performing some preliminary input compression and identification of core information.
 
\paragraph{Effect of context length for the abstractive step}
Long context models generally outperform their short context counterparts, with some exceptions. Long context models without an extractive step outperform short context models without an extractive step on all metrics, except BARTScore. When an extractive step is used, results vary as short context models show advantages on specific metrics. In other words, models with shorter input contexts benefit from input compression, as expected. Long context models without an extractive step generally outperform short context models with an extractive step across all metrics.

\subsection{Human evaluation}
\label{subsec: Human evaluation}

Human evaluation scores are in Table \ref{tab: Average human evaluation results}. Experts' individual scores and comments are in Appendix \ref{appendix: Human evaluation results}. 
Recall that the human evaluation was performed after selecting the optimal extractive model and fine-tuning all abstractive models. 
Overall, the expert evaluators preferred architectures that relied on a legal LM or a long context model in the extractive step. Indeed, the model that was preferred across all criteria was BART coupled with a LexLM extractor with a dependent compression ratio. The experts' comments suggested that this architecture did have shortcomings, but these were counterbalanced by other factors. For example, one expert noted that the summary correctly grasps the key points of the regulation, making it quite useful, despite the fact that is it incomplete and has shortcomings on fluency and coherence. 

Common criticisms of the summaries by the experts included excessive repetition in the case of some architectures, which severely decrease the quality of the produced summary. Furthermore, while some summaries may appear well-structured and readable, they fail to capture the essential points of the regulation or contain factual errors. 

A somewhat surprising outcome is that LLaMA-3 scores relatively poorly on coherence and fluency, compared to the best-performing model. It should be noted that the two evaluators diverged significantly in their scores for this model on these criteria (compare Tables~\ref{tab: Human evaluation results participant 1.} and \ref{tab: Human evaluation results participant 2.} in Appendix \ref{appendix: Human evaluation results}). Furthermore, as noted above, LLaMA was treated somewhat differently since it is the only decoder-only model. In particular, we subtracted 1500 tokens from the model's context length to accommodate the extractive summary; this too could have impacted results, though we adjusted the extractive summarisation process to ensure minimal truncation.

Despite the fact that this is a small-scale evaluation (a point we return to in Section~\ref{sec: Conclusion}), there are interesting divergences between expert judgments and the conclusions drawn based on the automatic metrics, an observation which is quite common in the NLG and summarisation literature 
\cite[cf.][]{Belz2006,reiter_structured_2018,celikyilmaz_evaluation_2021}. 

In particular, experts suggest that legal LMs help achieve more satisfactory summaries if used in the extractive step. On the other hand, both automatic and human evaluation suggest that BART is a competitive model for summarisation, especially if preceded by an extractive step. 


\section{Conclusion}
\label{sec: Conclusion}
In this paper, we focused on summarisation of long regulatory documents. Our findings indicate that while models with a longer context length do not benefit from extraction, an extractive step renders BART, an encoder-decoder architecture, highly competitive. A small-scaled evaluation with human experts confirms this finding. However, experts also indicate a preference for summaries relying on extraction with a domain-specific, legal language model. 

Future work should consider whether these findings are generalisable to other domains. Furthermore, a more extensive human evaluation is required to ensure that our findings are reliable. This is particularly crucial given that human expert judgments are not perfectly aligned with the outcomes of our metric-based evaluation, which echoes findings from other studies. A further possible research direction is to use a state-of-the-art LLM as an evaluator or `judge' for generated texts, a strategy which recent research suggests is increasingly viable \cite{liu_g-eval_2023,zheng_judging_2023}, though also one that requires some caution in view of results suggesting self-bias on the part of LLMs \cite{panickssery_llm_2024}, as well as lower reliability in comparison with expert judgment \cite{bavaresco_llms_2024}.



\bibliography{anthology,custom}
\clearpage
\appendix
\section{Further details on the method}
\label{appendix:Method}

\subsection{Derivation of $N$}
\label{appendix: Derivation of N}
The following is the derivation of Equation \ref{equation: calculate steps for fixed ratio}:
\begin{enumerate}
    \item The length of the intermediary summary $|E^j|$ after the first step is \( R \cdot |D| \). After the second step, it is \( R^2 \cdot |D| \) and so on. This implies that the length of the intermediary summary after $N$ steps is: 
\[|E^N|=R^N \cdot |D|\]
    \item Extractive steps are performed until the length of the intermediary summary is within the context length of the abstractive summarisation model, $K$:
\[ R^N \cdot |D| \leq K \]
    \item  To estimate $N$, take the logarithm on both sides:
\[ N \cdot \log(R) \leq \log\left(\frac{K}{|D|}\right) \]
    \item Then, solve for $N$:
\[ N \leq \frac{\log\left(\frac{K}{|D|}\right)}{\log(R)} \]
    \item $N$ is then rounded up to the highest integer. So, the formula for estimating the number of extractive steps $N$ needed before the final abstractive step can be taken is:
\begin{equation}
N = \left\lceil \frac{\log\left(\frac{K}{|D|}\right)}{\log(R)} \right\rceil
\end{equation}
\end{enumerate}

\subsection{Hyperparameter settings}
\label{appendix: Hyperparameter settings}

Table \ref{tab: Hyperparameters settings} summarises the hyperparameters used to finetune BART, T5, LongT5, Pegasus and PegasusX. 

\begin{table}[h]
  \begin{tabular}{ll}
    \hline
    Hyperparameter & Setting \\
    \hline
    Learning rate & 5$e^{-05}$ \\
    Epochs & 40 \\
    Effective batch size & 16 \\
    Warmup ratio & 0.1 \\
    Weight decay & 0.01 \\
    Early stopping patience & 5 \\
    Metric for best model & Validation loss \\
    Maximum generation length & 1500 \\ 
    \hline
  \end{tabular}
  \caption{Hyperparameter settings for BART, T5, LongT5, Pegasus and PegasusX.}
  \label{tab: Hyperparameters settings} 
\end{table}

\newpage
\subsection{Llama3 hyperparameter settings and training procedure}
\label{appendix: Llama3 hyperparameter settings}
Table \ref{tab: Llama3 settings} shows the hyperparameters used to finetuned Llama3 on the abstractive evaluation task.

\begin{table}[h]
  \begin{tabular}{ll}
    \hline
    Hyperparameter & Setting \\
    \hline
    Learning rate & $5e^{-05}$ \\
    Epochs & 10 \\
    Effective batch size & 16 \\
    Warmup ratio & 0.1 \\
    Weight decay & 0.01 \\
    Early stopping patience & - \\
    Metric for best model & - \\ 
    LoRA rank ($r$) & 8 \\
    LoRA alpha & 16 \\
    LoRA dropout & 0.1 \\
    Precision for frozen model weights & 4-bit NF\\
    Precision for low-rank matrices & bfloat16 \\
    Precision for calculations & bfloat16 \\
    Double Quantization & True\\
    \hline
  \end{tabular}
  \caption{\label{tab: Llama3 settings}
  Llama3 settings.}
\end{table}
Fully Sharded Data Parallel (FSDP) \cite{zhao2023pytorch} was used to fine-tune Llama3 with the \href{https://huggingface.co/docs/accelerate/usage_guides/fsdp}{Hugging Face implementation}. Due to issues when combining FSDP and QLoRA, the best-performing model could not be loaded, and early stopping patience and best model metric were not set. To mitigate overfitting, we used 10 epochs instead of 40, based on preliminary results indicating convergence between 4-20 epochs. For QLoRA, low-rank matrices were injected into the query, key, value matrices, and linear layers of Llama3, following settings from prior research \cite{raschka2023practical} \cite{hu2021lora}. To fine-tune Llama3, we combined the reference text and golden reference summary into a single sequence, providing Llama3 with the following input sequence:
\begin{center}
\begin{quote}
\texttt{Summarise the following text.}\\
\texttt{\#\#\# Text:} \\
\texttt{\{reference text\}}\\
\texttt{\#\#\# Summary:} \\
\texttt{\{golden reference summary\}}
\end{quote}
\label{prompt of Llama}
\end{center}
During prediction, no exemplary summary was given, allowing Llama3 to create a new summary.

\subsection{Evaluation metrics details}
\label{appendix: Evaluation metrics details}

We implemented ROUGE \cite{Lin2004} and BERTScore \cite{Zhang2020} using the \href{https://github.com/huggingface/evaluate/tree/main}{HuggingFace evaluate library}, comparing predictions against reference summaries using F-scores. For BERTScore, we employed the Longformer \cite{beltagy2020longformer} architecture for its long context length. BARTScore \cite{Yuan2021} was implemented with Stanford's \href{https://github.com/stanfordnlp/string2string}{string2string library}, using \href{https://huggingface.co/facebook/bart-large-cnn}{BART\cite{Lewis2020} fine-tuned} on the CNN/Daily Mail dataset \cite{Nallapati2016}. BARTScore calculates precision and recall based on log-likelihood, combined into an F-score, and is limited by BART's 1024-token context length. We used BLANC-help \cite{Vasilyev2020} from the \href{https://github.com/PrimerAI/blanc/tree/master}{BLANC package}, with a gap of two as this best correlates with human evaluation \cite{DBLP:journals/corr/abs-2010-06716}. BLANC, using BERT base \cite{Devlin2019}, is limited by its 512-token context.

\section{Human evaluation results}
\label{appendix: Human evaluation results}

Individual results for the two expert evaluations on each criterion are shown in Tables~\ref{tab: Human evaluation results participant 1.} and \ref{tab: Human evaluation results participant 2.}. These results are the basis for the averaged results in Section~\ref{subsec: Human evaluation} in the main paper. Below, we also summarise the main observations from the evaluators' comments on the summary outputs, for each architecture (architectures are numbered according to the order in the tables).

\paragraph{Architecture 1}
The evaluators indicated that the summary is not usable for readers without prior knowledge of the topic due to its incompleteness, factual mistakes, and inaccuracies. While it does touch upon the main principle of CBAM, some of the procedures and rules are described incorrectly.

\paragraph{Architecture 2}
The evaluators indicated that the summary is not usable for readers as it places information in the wrong place, describing background details in the ‘key points’ section instead of the main content of the regulation. Additionally, one evaluator mentions that the summary completely misses the main point of what CBAM is, despite the state information being mostly correct with only a few mistakes.

\paragraph{Architecture 3}
One evaluator indicates that the summary correctly grasps the key points of the regulation, making it quite useful. However, the evaluator noted that it is not fully complete and that the fluency and coherence of the sentences could be improved. Despite these shortcomings, the summary is considered a good starting point.

\paragraph{Architecture 4}
One evaluator noted that this summary is less flawed than that generated by Architecture 1 but is still unusable due to containing a significant amount of false information and incorrect words.

\paragraph{Architecture 5}
Both mentioned that the summary contains excessive repetitions. Although the summary starts well, its usability degrades as more repetitions are encountered. 

\paragraph{Architecture 6}
One evaluator states that the summary contains quite some useful information. However because the summary contains a lot of repetition, it becomes unusable.

\begin{table*}
\small
\centering
\hspace*{-2cm}
\begin{tabular}{llllccccc}
\hline
Architecture \# & Extr. model & Ratio & Abstr. model & FC & U & Acc & Fl & Coh \\
\hline
1 & RoBERTa & Dep. & BART & 1 & 1 & 1 & 1 & 1 \\
2 & - & NE & BART & 3 & 1 & 1 & \textbf{2} & 1 \\
3 & LexLM & Dep. & BART & \textbf{4} & \textbf{3} & \textbf{3} & \textbf{2} & \textbf{2} \\
4 & Longformer & Dep. & BART & 1 & 1 & 1 & 1 & 1 \\
5 & - & NE & PegasusX & \textbf{4} & 1 & 2 & 1 & 1 \\
6 & RoBERTa & Dep. & Llama3 & 3 & 1 & 2 & 1 & 1 \\
\hline
\end{tabular}
\caption{Human evaluation results participant 1. Dep: Dependent ration; NE: No extraction; FC: Factual Correctness; U: Usability; Acc: Accuracy; Fl: Fluency; Coh: Coherence}
\label{tab: Human evaluation results participant 1.}
\end{table*}

\begin{table*}
\small
\centering
\hspace*{-2cm}
\begin{tabular}{llllccccc}
\hline
Architecture \# & Extr. model & Ratio & Abstr. model & FC & U & Acc & Fl & Coh \\
\hline
1 & RoBERTa & Dep. & BART & 3 & 3 & 2 & 2 & 3 \\
2 & - & NE & BART & \textbf{4} & 1 & 3 & 4 & 2 \\
3 & LexLM & Dep. & BART & \textbf{4} & \textbf{4} & 3 & 4 & \textbf{4} \\
4 & Longformer & Dep. & BART & 3 & 3 & \textbf{4} & 2 & 3 \\
5 & - & NE & PegasusX & 3 & 1 & 3 & \textbf{5} & 1 \\
6 & RoBERTa & Dep. & Llama3 & 3 & 4 & 3 & 4 & 3 \\
\hline
\end{tabular}
\caption{Human evaluation results participant 2. Dep: Dependent ration; NE: No extraction; FC: Factual Correctness; U: Usability; Acc: Accuracy; Fl: Fluency; Coh: Coherence}
\label{tab: Human evaluation results participant 2.}
\end{table*}

\end{document}